# Chart-driven Connectionist Categorial Parsing of Spoken Korean

WonIl Lee
Geunbae Lee
Jong-Hyeok Lee
Department of Computer Science & Engineering
Pohang University of Science & Technology
San 31, Hyoja-Dong, Pohang, 790-784, Korea
bdragon@platon.postech.ac.kr

## Abstract

While most of the speech and natural language systems which were developed for English and other Indo-European languages neglect the morphological processing and integrate speech and natural language at the word level, for the agglutinative languages such as Korean and Japanese, the morphological processing plays a major role in the language processing since these languages have very complex morphological phenomena and relatively simple syntactic functionality. Obviously degenerated morphological processing limits the usable vocabulary size for the system and word-level dictionary results in exponential explosion in the number of dictionary entries. For the agglutinative languages, we need sub-word level integration which leaves rooms for general morphological processing.

In this paper, we developed a phoneme-level integration model of speech and linguistic processings through general morphological analysis for agglutinative languages and a efficient parsing scheme for that integration.

Korean is modeled lexically based on the categorial grammar formalism with unordered argument and suppressed category extensions, and chart-driven connectionist parsing method is introduced.

## 1 Introduction

Spoken language processing challenges for integration of speech recognition into natural language processing, and must deal with multi-level knowledge sources from signal level to symbol level. The multi-level knowledge integration and handling increase the technical difficulty of both the speech and the natural language processing. In the speech recognition side, the recognition must be at phoneme-level for large vocabulary continuous speech, and the speech recognition module must provide right level of outputs to the natural language module in the form of not single solution but many alternatives of solution hypotheses. The n-best list (Chow & Schwartz 1989), word-graph (Oerder & Ney 1993), and word-lattice (Murveit et al. 1993) techniques are mostly used in this purpose. The speech recognition module can also ask the linguistic scores from the language processing module in a more tightly coupled bottom-up/top-down hybrid integration scheme (Paul 1989). In the natural language side, the insertion, deletion, and substitution errors of continuous speech must be compensated by robust parsing and partial parsing techniques, e.g. (Baggia & Rullent 1993). Often the spoken languages are ungrammatical, fragmentary, and contain non-fluencies and speech repairs, and must be processed incrementally under the time constraints (Menzel 1994).

Most of the speech and natural language systems which were developed for English and other Indo-European languages neglect the morphological processing, and integrate speech and natural language at the word level (Bates et al. 1993; Agnas et al. 1994). Often these systems employ a pronunciation dictionary for speech recognition and independent dictionaries for natural language processing. However, for the agglutinative languages such as Korean and Japanese, the morphological processing plays a major role in the language processing since these languages have very complex morphological phenomena and relatively simple syntactic functionality. Unfortunately even the Japanese researchers apply degenerated morphological techniques for the spoken Japanese processing (Hanazawa et al. 1990; Sawai 1991). Obviously degenerated morphological processing limits the usable vocabulary size of the system, and word-level dictionary results in exponential explosion in the number of dictionary entries. For the agglutinative languages, we need sub-word level integration which leaves rooms for general morphological processing.

In this paper, we propose a parsing scheme for spoken Korean, called $C^3$ parsing, which is integrated with the speech recognition in phoneme-level.

## 2 Overall parser architecture

We call our parsing scheme as $C^3$ parsing which stands for chart-driven connectionist categorial parsing. Korean is modeled by an extended categorial grammar and parsed by a connectionist method which is controlled by the charts.

## 3 Extension of Categorial Grammar
### 3.1 Directional categorial grammar

The directional categorial grammar naturally represents some Korean morphemes, such as postpositions(noun-endings and verb-endings), adverbs, and pre-nouns. A directional categorial grammar(Uszkoreit 1986; Zeevat 1988) is defined as an ordered quintuple G = <V, C, S, R, f>, where

- V: the vocabulary set,
- C: a finite set of basic categories which generates a full set C' of categories by the recursive application of the following category formation rules:
  if a ∈ C, then a ∈ C' and
  if a ∈ C' and b ∈ C', then a/b ∈ C' and a\b ∈ C',
- S: the category for sentences,
- R: a set of functional application rules such as
  left cancelation
    a b\a ⇒ b
  right cancelation
    b/a a ⇒ b
- f: an assignment function of elements of V to subsets of C'.

  Here comes a few example assignments:
  say(new)                np/np
  pha-il(file)             np
  tul(-s)                  np\np
  ul(objective case marker) np[obj]\np.

These assignments can be used to build a np[obj] structure for the eojeol "pha-il-tul-ul" as shown in figure 1.

### 3.2 Unordered arguments extension

Korean has relatively free word order compared to SOV languages (Lee, Lee, & Lee 1994). For example, the sentence "ku-ka sa-kwa-lul mek-nun-ta" (which means "he eats an apple") can be written as "sa-kwa-lul ku-ka mek-nun-ta". This word-order variations can't be modeled by the pure directional CG and We extend the category formation and functional application rules to deal with this phenomena:

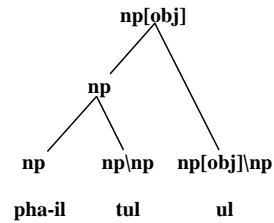

Figure 1: "pha-il-tul-ul" parsed by the directional CG.

- if a ∈ C, then a ∈ C'
- if a ∈ C', and S ⊂ C', then a/S ∈ C' and a\S ∈ C'
- left cancellation
  $a_i$ b\{$a_1$,...,$a_n$} ⇒ b\{$a_1$,...,$a_{i-1}$,$a_{i+1}$,...,$a_n$}
- right cancellation
  b/{$a_1$,...,$a_n$} $a_i$ ⇒ b/{$a_1$,...,$a_{i-1}$,$a_{i+1}$,...,$a_n$}

The sentences "ku-ka sa-kwa-lul mek-ess-ta" and "sa-kwa-lul ku-ka mek-ess-ta" can be parsed by the following category assignments (in figure 2):

  ku(he)                    $np$
  ka(subj. case-marker)     $np[subj]\backslash np$
  sa-kwa(apple)             $np$
  lul(obj. case-marker)     $np[obj]\backslash np$
  mek(eat)                  $s\backslash\{np[subj], np[obj]\}$
  nun-ta(declaritive modal) $s[tDEC]\backslash(s\backslash \$X)$

### 3.3 Suppressed categories extension

After a few experiments with the directional CG with unordered argument extension, we found so much structural ambiguities resulted even from simple Korean sentences. For example, "say pha-il-tul"(new files) is parsed in two ways(figure 3).

These abmiguities are caused by the fact that the categorial assignments we used to model Korean were too simple and had no order between rule applications. Here, we suggest a way to inhibit figure 3 (a) through suppressed and activator categories. A suppressed category is a category with | instead of / or \, which can be changed to a category with / or \ by an activator category. An activator category is a category with a suppressed category as its argument and an ordinary category as its result. Following categorial assignments shows suppressed and activator categories:

  say(new)                   $np/np$
  pha-il(file)                $np|$
  tul(plural suffix)          $np\backslash(np|)$
  ul(objective case marker)  $np[obj]\backslash np$.

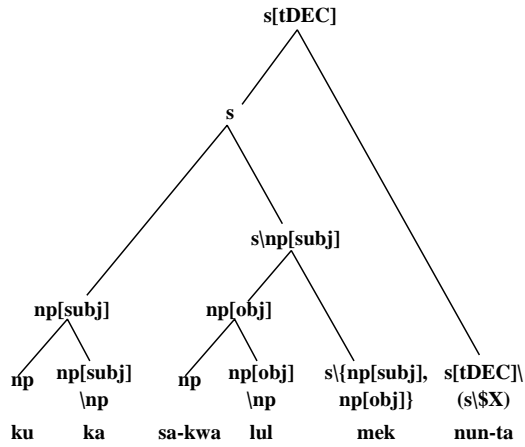

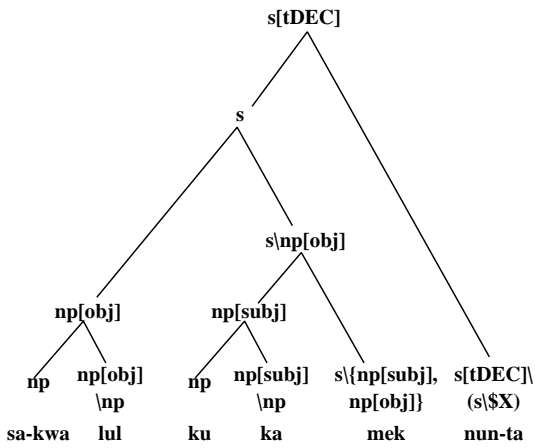

Figure 2: Word-order independant parse

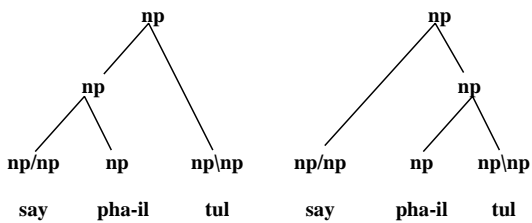

Figure 3: Two alternative parses for "say pha-il-tul", generated by CG

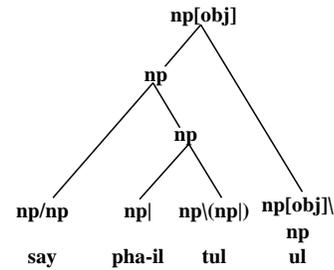

Figure 4: Correct parses for "say pha-il-tul" generated by CG with unordered arguments, and suppressed and activator categories.

The np| is a suppressed category and prevented from being combined with the "say" and forces the noun and the suffix to combine first. The np\(np|) is an activator category whose argument is a suppressed category and result is an ordinary category. Whether a morpheme gets an ordinary, a suppressed, or an activator category is determined by the morphological conditions. A noun followed by a suffix gets a suppressed category and a suffix followed by a noun-ending gets an activator category. This assignment is handled by the morphological classification information and morpheme connectivity matrix.

Figure 4 shows the parse for the "say pha-il-tul" using suppressed and activator categories.

These two classes of categories lexically models Korean noun structures and predicate structures and can be used for other agglunative languages.

## 4 $C^3$ parsing

The syntax analysis is performed by interactive relaxation (spreading activation) parsing on the categorial grammar where the position of the functional applications are controlled by a triangular table. The original interactive relaxation parsing (Howells 1988) was extended to provide efficient constituent searching and expectation generation through positional information provided by categorical grammar and triangular table (Lee & Lee 1995 in press) and to parse the input morpheme lattice at once. The interactive relaxation process consists of the following three steps that are repetitively executed: 1) add nodes, 2) spread activation, and 3) decay.

- add nodes

    Grammar nodes (syntactic categories from the dictionary) are added for each sense of the morphemes when the parsing begins. A grammar node which has more activation than the predefined threshold $\Theta$ generates new nodes in

the proper positions (to be discussed shortly). The newly generated nodes search for the constituents (expectations) which are in the appropriate table positions, and are of proper function applicable categories.

- spread activation

  The bottom-up spreading activation is as follows:
  $$n \times \rho \times a \times \frac{a_i^2}{\sum_{j=1}^n a_j^2}$$
  where predefined portion $\rho$ of total activation $a$ is passed upward to the node with activation $a_i$ among the $n$ parents each with node activation $a_j$. In other words, the node with large activation gets more and more activation, and it gives an inhibition effects without explicit inhibitory links (Reggia 1987).

  The top-down spreading activation uniformly distributes:
  $$\rho' \times a$$
  among the children where $\rho'$ is predefined portion of the source activation $a$.

- decay

  The node's activation is decayed with time. The node with less constituents than needed gets penalties plus decays:
  $$a \times (1 - d) \times \frac{Ca}{Cr}$$
  where $a$ is an activation value, $d$ is a decay ratio, and Ca, Cr is the actual and required constituents. After the decay, the node with less activation than the predefined threshold $\Phi$ is removed from the table.

The node generation and constituent search positions are controlled by the triangular table. When the node a(i,j) acts as an argument, it generates node only in the position (k,j) where $1 < k < j$, and the generated node searches for the constituents (functors) only in the position (k,i-1). Or when the node is generated in the position (i,k) where $j < k < number - of - morphemes$, it searches for the position (j+1,k) for its constituents. When the node acts as a functor, the same position restrictions also apply for the node generation and the argument searching. The position control combined with the interactive relaxation guarantees an efficient, lexically oriented, and robust syntax analysis of spoken languages.

## 5 Spoken Korean morphological analysis

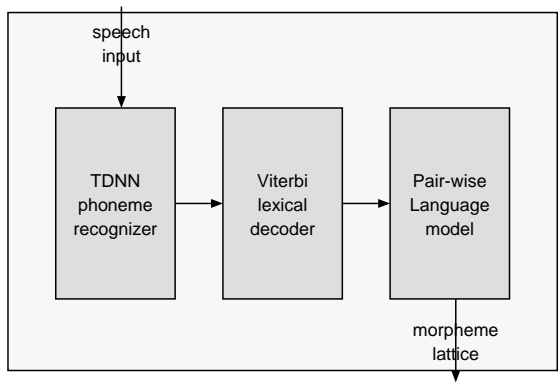

Figure 5: The TDNN-based speech recognizer

Figure 5 shows the architecture of our TDNN-based speech recognizer. The TDNN(time-delayed neural network)-based phoneme recognizer gvies a sequence of phonemes for the input speech and this phoneme sequence is decoded by the viterbi lexical decoder. Tree-structured phoneme-sequence-to-morpheme dictionary is used in the lexical decoding phase and a morpheme lattice is extracted. This lattice is filtered by the pairwise language model. The language model checks each adjacent pair of morphemes in the lattice whether they are connectable morphologically and phonologically. Then the remaining morpheme-lattice is given to the parser.

## 6 Experimental results

Through this section, we focus on testing the effect of extensions on the categorial grammar and the reliability of the parser. We did 4 simulated speech recognition experiments using extended categorial grammars:

- Unordered argument base-line(UAB)
- Unordered argument phoneme lattice(UAP)
- Unordered argument + suppressed category base-line(UA+SCB)
- Unordered argument + suppressed category phoneme lattice(UA+SCP)

Base-line experiments are to see the complexity of the corpus and the modeling ability of the extended grammar (figure 6). Single correct phoneme sequence for each sentence is given to the parser.

Phoneme-lattice experiments are to see the parsing and selecting ability of the parser when more than one candidates are given for each phoneme which increase the ambiguity (figure 7). 10 phoneme lattices are generated for each sentence. For each correct phoneme, about 2.2 candidates are generated along the confusion matrix between Korean phonemes. Each phoneme lattice was made to contain at least one correct recognition result, so the phoneme recognition performance is assumed to be perfect in the lattice form.

|  | Morphological analysis (morpheme lattice) | Syntactic analysis (best 1 parse) |
|---|---|---|
| UAB | 100 % (33/33) | 60.6 % (20/33) |
| UA+SCB | 100 % (33/33) | 78.8 % (26/33) |

Figure 6: The morphological and syntactic analysis results for correct phoneme sequences.

|  | Morphological analysis (morpheme lattice) | Syntactic analysis (best 1 parse) |
|---|---|---|
| UAP | 100 % (330/330) | 35.5 % (117/330) |
| UA+SCP | 100 % (330/330) | 61.8 % (240/330) |

Figure 7: The morphological and syntactic analysis results for the artificially made phoneme lattices.

Totally 33 sentences from the UNIX natural language interface (Lee & Lee 1995 in press) are used in the test. The sentences (translated in english) includes :

- Show me the users who are logging in this machine.
- Show me the users who are in idle more than an hour.
- Send them a mail.
- Let me know how much disk space "bdragon" uses.
- Change directory to the home directory of "bdragon".

We built two morpheme-level phonetic dictionaries for about 1000 morphemes, one with the unordered-argument categorial information and the other with the unordered-argument plus suppressed categorial information. For the syntactic level interactive relaxation, we used the following parameters (which are experimentally determined): upward propagation portion $\rho$ 0.05, downward propagation portion $\rho$' 0.03, decay ratio $d$ 0.87, the node generation threshold $\Theta$ 0.51, and the node removal threshold $\Phi$ 0.066. Using the activation level of each parse tree, one best parse is selected.

The morphological analysis was perfect in all 4 experiments as shown in the tables. Since the phoneme lattice was made to contain at least one correct phoneme recognition result, the morphological analysis must be perfect as long as the morpheme is enrolled in the dictionary and the connectivity information can cover all the morpheme combinations. This was possible due to the small number of tested sentences (33 sentences). This results verify that most of the morphological analysis errors from real speech input are actually propagated from the phoneme recognition errors as discussed before.

UAB compared to UA+SCB shows that the lack of noun or predicate structure modeling results in extra structural ambiguities and this differentiated the UAP and UA+SCP more. In UA+SCB we can see an improved performance due to the noun or predicate structure modeling through suppressed categories.

Through UAP and UA+SCP we faced 20 to 30 percent performance down due to the multiple candidates in the phoneme lattices that results in more ambiguous morpheme lattices, and finally ambiguous syntactic trees. These failures can be reduced if we generate n-best parse trees, and let the semantic processing module select the correct ones as is usually done in most of the probabilistic parsing schemes(Charniak 1994).

## 7 Concluding remarks

We developed a phoneme-level integration model of speech and linguistic processings through general morphological analysis for agglutinative languages and a efficient parsing scheme for that integration. Through unordered argument and suppressed category extensions, we modeled Korean lexically based on categorial grammar formalism and the justifications for that extensions are shown by experiments. The chart-driven interactive relaxation parsing and extended categorial grammar showed robust handling of word-order variations and complex word structures in Korean. The performance of the system (78.8% and 61.8%) can be improved if we use n-best strategy coupled with the semantic processing.

We think the system can be extended to other agglutinative languages such as Japanese, Finish, and Turkish, and the languages that have complex morphological phenomena such as German and Dutch, since phonological and orthographic rules, the word order and the word structures are modeled declaratively only through the dictionary.